\documentclass[letterpaper, 10 pt, conference]{ieeeconf}
\IEEEoverridecommandlockouts                              

\usepackage{amsmath,amsfonts}
\usepackage{algorithmic}
\usepackage{algorithm}
\usepackage{array}
\usepackage{textcomp}
\usepackage{stfloats}
\usepackage{url}
\usepackage{verbatim}
\usepackage{graphicx}
\usepackage{cite}
\usepackage{comment}
\usepackage{threeparttable}
\newtheorem{lemma}{Lemma}
\usepackage{booktabs} 
\usepackage{siunitx}  
\usepackage{multirow}
\newcommand{\Snq}{{$\boldsymbol{\Sigma}_{\mathbf{n}, \mathbf{q}}\ $}}

\title{\LARGE \bf
C\textsuperscript{3}P-VoxelMap: Compact, Cumulative and Coalescible Probabilistic Voxel Mapping}

\author{
        Xu Yang, Wenhao Li, Qijie Ge, Lulu Suo, Weijie Tang, Zhengyu Wei, Longxiang Huang, and Bo Wang
\thanks{All authors are with Deptrum Ltd. 
Wenhao Li and Qijie Ge contributed equally to this work. 
Corresponding author: Xu Yang \tt\small{xu.yang@deptrum.com}}%
}

\begin{document}
\maketitle
\thispagestyle{empty}
\pagestyle{empty}



\begin{abstract}
This work presents a compact, cumulative, and coalescible probabilistic voxel mapping method to enhance performance, accuracy, and memory efficiency in LiDAR odometry.
Probabilistic voxel mapping requires storing past point clouds and re-iterating them to update the uncertainty at every iteration, which consumes large memory space and CPU cycles.
To solve this problem, we propose a two-fold strategy.
First, we introduce a compact point-free representation for probabilistic voxels and derive a cumulative update of the planar uncertainty without caching original point clouds.
Our voxel structure only keeps track of a predetermined set of statistics for points that lie inside it.
This method reduces the runtime complexity from $O(MN)$ to $O(N)$ and the space complexity from $O(N)$ to $O(1)$ where $M$ is the number of iterations and $N$ is the number of points.
Second, to further minimize memory usage and enhance mapping accuracy, we provide a strategy to dynamically merge voxels associated with the same physical planes by taking advantage of the geometric features in the real world.
Rather than constantly scanning for these coalescible voxels at every iteration, our merging strategy accumulates voxels in a locality-sensitive hash and triggers merging lazily.
On-demand merging reduces memory footprint with minimal computational overhead and improves localization accuracy thanks to cross-voxel denoising.
Experiments exhibit 20\% higher accuracy, 20\% faster performance, and 70\% lower memory consumption than the state-of-the-art.
\end{abstract}


\section{Introduction}\label{Introduction}
LiDAR SLAM has been extensively studied in the past decade thanks to the wide availability of depth sensors.
The direct and precise depth acquisition facilitates pose estimation and map reconstruction.
Numerous works contribute to the advancement of LiDAR SLAM \cite{LEGO-LOAM, DLIO, VOXELMAP, PLC-SLAM, FASTER-LIO, FAST-LIO2, SSL-SLAM}.
These works can be broadly divided into direct and indirect methods based on how they use raw point clouds.

Indirect methods explicitly extract geometric features like straight lines or planes from point clouds and perform localization and mapping with respect to the extracted features\cite{LEGO-LOAM, PLC-SLAM, SSL-SLAM}.
Although effective in creating sparse maps, feature extraction and correspondence association are computationally intensive and heavily rely on sensor-specific patterns like circular LiDAR scans.

In contrast, direct methods obviate the need for feature extraction and directly register raw points to the map\cite{DLIO, FAST-LIO2}.
These approaches leverage all available raw points for mapping and create dense maps.
Consequently, there is a growing demand for efficient map organization in real-time LiDAR SLAM, particularly on resource-constrained platforms like embedded devices.

Direct methods usually adopt one of the two kinds of map representations for point registration: irregular tree structures or regular voxel maps.
K-d tree partitions the space into blocks of dynamic sizes.
To locate a plane for pose estimation, a tree traverse searches for the nearest neighbors \cite{FAST-LIO2, DLIO}.
The performance of traversal may degrade when the tree is heavily unbalanced.
Tree re-balancing and maintenance are complicated and often cause unexpected global pauses for real-time tasks.
In contrast, hash-based voxel maps are more effective and universal.
Voxel map partitions the space into blocks of fixed size and locates them with a hash function.
Recent works \cite{FASTER-LIO, VOXELMAP} have demonstrated the outstanding efficiency of voxel maps.
Given these advantages, voxel maps still face several challenges:
\begin{figure}[t]
    \centering
    \includegraphics[width=0.9\linewidth]{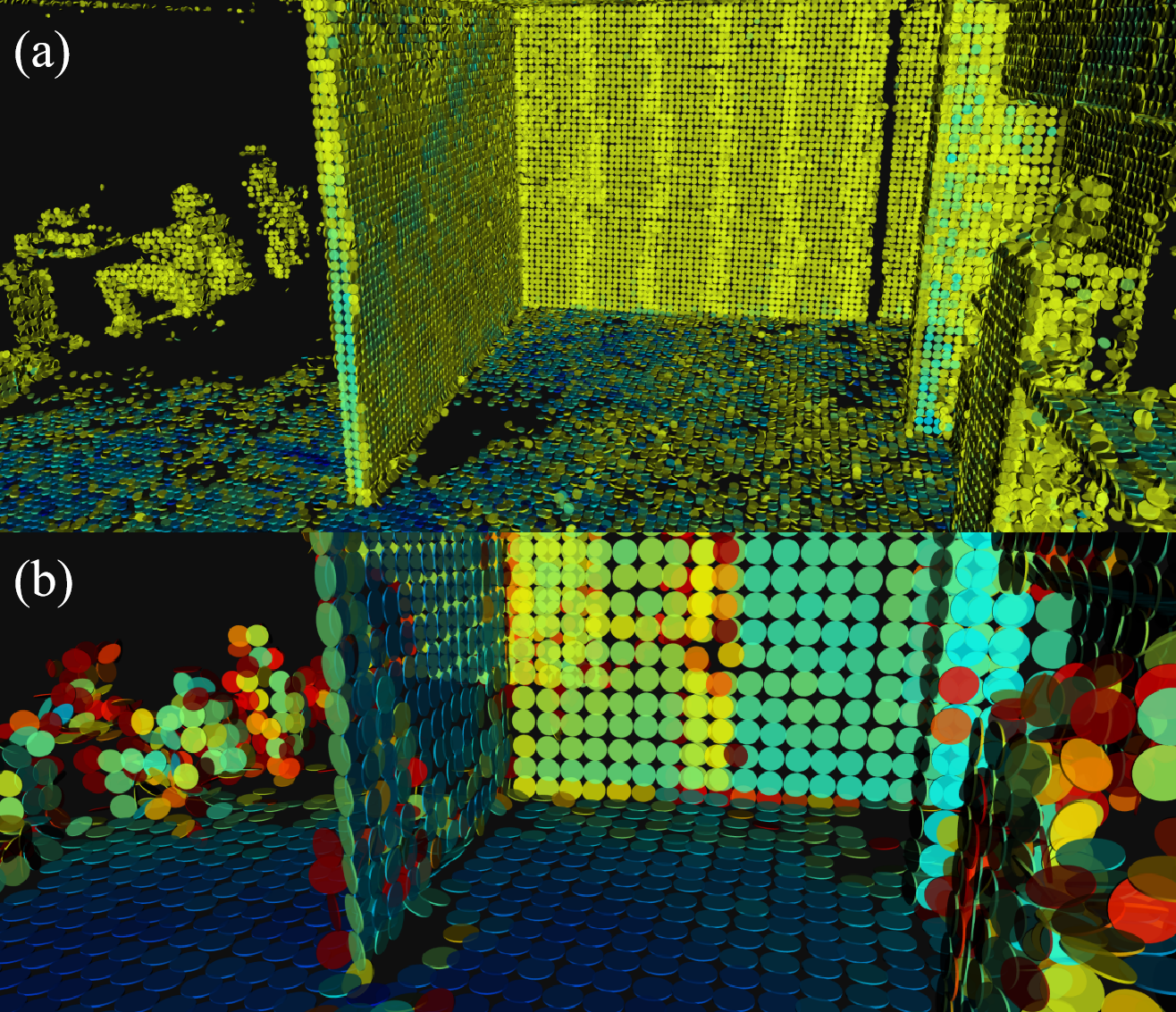}
    \caption{Bias-variance tradeoff of voxel maps. (a) Small voxels capture details of the environment but are susceptible to noise, i.e., large variance (note the wiggling planes on the walls and the ground). (b) Large voxels are more resistant to noise but might lead to approximation bias (note the thick wall is over-smoothed into a plane).}
    \label{fig:bias-var}
\end{figure}
\begin{enumerate}
\item {Probabilistic voxel mapping requires keeping all points to update the map because an iteration on all points is required to recompute the uncertainty of voxel planes\cite{VOXELMAP}.
This non-cumulative process is both time-consuming and memory-inefficient.}
\item{A regular voxel structure cannot capture large geometric features, e.g., walls, floors, ceilings, in the real world.
This inflexibility leads to a duplication in representing a single large feature with many voxels.}
\item{Bias-variance tradeoff on voxel sizes.
Large voxels excel at collecting a substantial amount of points to reduce variance, but they tend to oversmooth the environment, resulting in a representation bias.
Conversely, small voxels excel at capturing map details, i.e., lower bias, but are susceptible to noise, i.e., higher variance, especially when the point cloud is sparse.
(see Fig. \ref{fig:bias-var})
}
\end{enumerate}

To address the above challenges, we propose C\textsuperscript{3}P-VoxelMap with the following contributions:
\begin{enumerate}
\item{We derive a compact point-free representation of plane uncertainty and a cumulative update scheme for each voxel.
Our approach eliminates the recalculation of Jacobians for plane parameters w.r.t. each point and reduces the computation workload substantially.
Moreover, the memory consumption of each voxel is independent of the number of points therein, resulting in significant memory savings.}
\item{We introduce a voxel merging strategy to dynamically adapt to large planar features in the real world.
With voxel merging, a large planar feature only requires one or a few voxels to represent.
Moreover, voxel merging improves mapping accuracy by estimating a single large feature with points from multiple voxels.}
\item{We propose an on-demand merging strategy with a locality-sensitive hash that triggers a merge operation only when enough coalescible voxels are aggregated.
This strategy realizes voxel merging with minimal overhead and avoids brute-force search.}
\end{enumerate}

\section{Related Works}
This section briefly reviews the related works on direct and voxel-based mapping methods in recent years and details the difference with our proposed C\textsuperscript{3}P-VoxelMap. 

Probabilistic voxel mapping originates from NDT \cite{NDT} and has evolved into many variations \cite{3D-NDT, LITAMIN2, VOXELMAP}.
\cite{3D-NDT} first proposed to model voxelized point clouds with Gaussian distributions and minimize the Mahalanobis distance between source points and the target voxel distributions, leading to an efficient inquiry and registration.
LiTAMIN\cite{LITAMIN} incorporates NDT's voxelization in the target point set, which is usually large in size while adopting a k-d tree for nearest neighbor search (NNS) in the source point set.
LiTAMIN2\cite{LITAMIN2} extends LiTAMIN by applying KL-divergence to measure the distribution-to-distribution distance. 

Different from Fast-LIO2\cite{FAST-LIO2} that uses NNS of a k-d tree, Faster-LIO\cite{FASTER-LIO} points out that the strict NNS is unnecessary for LIO and utilizes the voxel structure with approximate NNS.
To manage the unbounded growth of the map, incremental voxel pruning removes voxels that haven't been recently used.
Consequently, Faster-LIO achieves similar accuracy as Fast-LIO2\cite{FAST-LIO2} while being more efficient in computation and memory consumption.

In \cite{FAST-LIO2} and \cite{FASTER-LIO}, local planes are treated as deterministic features.
That is, the system only estimates the pose of a plane but not its uncertainty during state estimation.
However, the pose of a plane is regressed from points from multiple LiDAR scans and is estimated in the global frame, thus, the uncertainty raised from both LiDAR measurements and ego-pose estimation contributes to the latent plane distribution.
To address this issue, VoxelMap\cite{VOXELMAP} proposes a probabilistic plane representation and explicitly parameterizes the plane as a multivariate function of all points.
The uncertainty of a plane is jointly determined by the points and the ego-poses.

Probabilistic plane representation has demonstrated its effectiveness in improving mapping accuracy\cite{VOXELMAP}.
However, to update a plane with newly observed points, all past points are required to calculate the new uncertainty, which creates a big burden for both memory and computation.
To enhance memory efficiency, VoxelMap++\cite{VOXELMAP++} replaces the 6-DoF plane parameters with a 3-DoF representation.
Since each voxel contains only a single plane but hundreds of points, the memory usage is dominated by the points therein.
Thus, the space savings from 3-DoF representation are limited.
In this work, a compact representation and a cumulative plane update are proposed, eliminating the storage of past points and repeated computation.

To overcome the inability to adapt to large geometric features, \cite{BALM} and \cite{VOXELMAP} utilize a coarse-to-fine voxel hierarchy.
A regular voxel hash map is used for the coarse level, and for finer granularity, each voxel is further subdivided into sub-voxels that are indexed by an octree.
Representing larger or irregular features in this voxel hierarchy is still unmanageable.
\cite{VOXELMAP++} proposes to merge voxels with similar plane parameters.
This design shares some similarities with our on-demand merging, but there exist several key differences.
First, instead of searching for coalescible voxels constantly with union-find \cite{VOXELMAP++}, our on-demand merging is triggered only when enough mergeable voxels are gathered in the locality-sensitive hash. Therefore, the voxel merging overhead is much smaller.
Second, our merging strategy incorporates cross-voxel updates of probabilistic planes with respect to the uncertainty from all voxels. 

In sum, \cite{FAST-LIO2, FASTER-LIO, VOXELMAP, VOXELMAP++} are the most relevant works to ours. 
The differences among these methods are listed in Table \ref{method_comparison}.
In terms of the state estimation, all the above methods are developed under the Iterative Error State Kalman filter (IESKF).
Since this work mainly focuses on how to manage the voxel map compactly, we keep the state estimation algorithm the same. 
\cite{FAST-LIO2} gives a detailed introduction to IESKF.
\begin{table}[h]
\centering
\caption{Comparison of Methods}
\label{method_comparison}
\begin{threeparttable}
\begin{tabular}{cccc}
\toprule
     &  Map Structure & Plane Rep. & Plane Update \\
\midrule
  Fast-LIO2 & Incremental K-d tree & Det. & No update \\
  Faster-LIO &  Incremental Voxel & Det. & No update \\
  VoxelMap &  Adaptive Voxel & Prob. & Non-cumulative\\
  VoxelMap++ & Mergeable Voxel & Prob. & Non-cumulative \\
    Ours &  C\textsuperscript{3}P Voxel & Prob. & Cumulative \\
\bottomrule
\end{tabular}
\begin{tablenotes}
\item {Det. is short for Deterministic.}
\item {Prob. is short for Probabilistic.}
\end{tablenotes}
\end{threeparttable}
\end{table}

\section{System Overview}
C\textsuperscript{3}P-VoxelMap incorporates a coalescible voxel map with probabilistic plane representation and optimizes the estimation with IESKF. 
The voxel map is a 3D grid of the space where
each voxel stores a probabilistic plane extracted from point clouds aggregated across multiple frames. 
The uncertainty of a plane is represented by a covariance matrix, jointly determined by the noise of all past points and the associated camera poses.

IESKF optimizes probabilistic plane configurations and camera poses by minimizing the point-to-plane distance.
This probabilistic representation provides a more accurate point-to-plane measurement model than deterministic representations, as it accounts for the plane's uncertainty \cite{VOXELMAP}.
C\textsuperscript{3}P-VoxelMap enables cumulative update to the covariance of probabilistic planes without storing and re-iterating on past point clouds (see Section \ref{sec:Cumulative Probabilistic Update}).

To merge voxels representing the same physical planes on demand,
voxels are hashed into buckets based on their locations and plane orientations. 
When a bucket accumulates sufficient voxels, small planes are merged into a large one. 
The uncertainty of the merged plane is updated using the same cumulative method as for individual voxel planes. 
Each voxel subsequently adopts the merged plane as its feature in the filtering process
(see Section \ref{sec:on-demand-merging}).

\section{Cumulative Probabilistic Update}
\label{sec:Cumulative Probabilistic Update}

\subsection{Probabilistic Plane Representation}
\label{PPR}
Following the formulation in probabilistic voxel mapping \cite{VOXELMAP}, our system employs planes as the feature in the voxel map, which exist ubiquitously in the real world.
A probabilistic plane is composed of a normal vector $\mathbf{\mathbf{n}}$, a center point $\mathbf{q}$, and a covariance matrix \Snq denoting the uncertainty.
The uncertainty model of a voxel plane considers both sensor noise and the estimated pose noise in the world frame. 

Considering a point $\mathbf{p}$ as a 3-dimensional random variable sampled from a plane, the parameter of the plane can be formulated as a multi-variate function $f$ of all points:
\begin{equation}
\begin{aligned}
\label{eq2}
\left[\mathbf{n}^{g t}, \mathbf{q}^{g t}\right]^T
&=f\left( \mathbf{p}_1+\boldsymbol{\delta}_{{\mathbf{p}_1}}, \mathbf{p}_2+\boldsymbol{\delta}_{{\mathbf{p}_2}}, \ldots, \mathbf{p}_N+\boldsymbol{\delta}_{{\mathbf{p}_N}}\right) \\
&\approx[\mathbf{n}, \mathbf{q}]^T+\sum_{i=1}^N \frac{\partial \mathbf{f}}{\partial \mathbf{p}_i} \boldsymbol{\delta}_{{\mathbf{p}_i}} \\
&
\end{aligned}
\end{equation}
where $\left[\mathbf{n}^{gt},\mathbf{q}^{gt}\right]$ denotes the ground truth plane, and $\boldsymbol{\delta}_{\mathbf{p}_i}$ denotes the Gaussian distributed noise of the $i$-th point. Here, function $f$ first computes the mean $\mathbf{q}$ and the covariance matrix $\mathbf{A}$ from the point clouds:
\begin{equation}
    \mathbf{q}=\frac{1}{N}\sum_{i=1}^N{\mathbf{p}_i} \qquad     \mathbf{A}=\frac{1}{N}\sum_{i=1}^N{\mathbf{p}_i\mathbf{p}_i^T-\mathbf{q}\mathbf{q}^T}
\end{equation}
then followed a singular value decomposition (SVD) of matrix $\mathbf{A}$ to obtain eigenvalues and eigenvectors:
\begin{equation}
    \mathbf{A}=\mathbf{U}\operatorname{diag}(\lambda_1, \lambda_2, \lambda_3)\mathbf{U}^T \qquad \mathbf{U}=\left[\mathbf{u}_1, \mathbf{u}_2, \mathbf{u}_3\right]
\end{equation}
where $\lambda_1$, $\lambda_2$ and $\lambda_3$ are singular values in a descend order, $\mathbf{u}_1$, $\mathbf{u}_2$ and $\mathbf{u}_3$ are singular vectors. 
The normal $\mathbf{n}$ of the voxel plane is simply obtained by
\begin{equation}
\label{eq:normal}
    \mathbf{n}=\mathbf{u}_3
\end{equation}
The Jocabian of $\mathbf{n}$, $\mathbf{q}$ w.r.t. each point $\mathbf{p}_i$ in  (\ref{eq2}) is computed as follows\cite{VOXELMAP}:
\begin{equation}
\label{eq:partial}
    \frac{\partial \mathbf{n}}{\partial \mathbf{p}_i}=\mathbf{U}\left[\begin{array}{c}
\left( \mathbf{p}_i-\mathbf{q}\right)^T\mathbf{F}_{1} \\
\left( \mathbf{p}_i-\mathbf{q}\right)^T\mathbf{F}_{2} \\
\left( \mathbf{p}_i-\mathbf{q}\right)^T\mathbf{F}_{3}
\end{array}\right],\frac{\partial \mathbf{q}}{\partial {\mathbf{p}_i}}=\operatorname{diag}\left(\frac{1}{N}, \frac{1}{N}, \frac{1}{N}\right)
\end{equation}
\begin{equation}
\label{eq3-1}
\mathbf{F}_{m}=\left\{\begin{array}{cc}
\frac{1}{N\left(\lambda_3-\lambda_m\right)}\left(\mathbf{u}_m \mathbf{n}^T+\mathbf{n} \mathbf{u}_m^T\right) & , m \neq 3, \\
\mathbf{0}_{1 \times 3} & , m=3 .
\end{array}\right.
\end{equation}
Thus, the plane uncertainty is a linear combination of all $\boldsymbol{\Sigma_{\mathbf{p}_i}}$:
\begin{equation}
\label{eq5}
\boldsymbol{\Sigma}_{\mathbf{n}, \mathbf{q}}=\sum_{i=1}^N \frac{\partial \mathbf{f}}{\partial \mathbf{p}_i} \boldsymbol{\Sigma_{{\mathbf{p}_i}}}{\frac{\partial \mathbf{f}}{\partial \mathbf{p}_i}}^T, \frac{\partial \mathbf{f}}{\partial \mathbf{p}_i}=\left[\frac{\partial \mathbf{n}}{\partial \mathbf{p}_i}, \frac{\partial \mathbf{q}}{\partial \mathbf{p}_i}\right]
\end{equation}
where $\boldsymbol{\Sigma_{\mathbf{p}_i}}\in\mathbb{R}^{3\times3}$ denotes the covariance matrix of the $i$-th point in the world coordinate frame.

\subsection{Cumulative Update}
\label{Cumulative}
Upon the arrival of new points, both the pose $\mathbf{\mathbf{n}}$, $\mathbf{q}$, and uncertainty \Snq of a voxel plane need to be updated.
Updating $\mathbf{q}$ cumulatively w.r.t. a new point $\mathbf{p}$ is straightforward.
\begin{equation}
    \mathbf{q}^{\prime} =\frac{N}{N+1}\mathbf{q}+\frac{1}{N+1}\mathbf{p}
\end{equation}

The update of $\mathbf{n}$ is based on an cumulative update of $\mathbf{A}$
\begin{equation}
\label{eqA}
\mathbf{A}^\prime=\frac{N}{N+1}(\mathbf{A}+\mathbf{q}\mathbf{q}^T)+\frac{1}{N+1}\mathbf{p}\mathbf{p}^T - \mathbf{q}^\prime\mathbf{q^\prime}^T
\end{equation}
The normal vector $\mathbf{n}^\prime$ is the third singular vector of the covariance matrix $\mathbf{A}^\prime\in\mathbb{R}^{3\times3}$ under singular value decomposition.

However, the cumulative update of \Snq is non-trivial because this update triggers a re-computation of the Jacobian of $\mathbf{n}$, $\mathbf{q}$ w.r.t. $ \mathbf{p}_i$.
As shown in  (\ref{eq5}), the re-computation involves every point that falls into a voxel in the past.
The re-computation is not only time-consuming but also memory-inefficient because all past points must be stored and re-iterated.
We derive a new algorithm that makes cumulative updates possible and avoids the storage of past points.

By expanding  (\ref{eq5}) we have
\begin{equation}
\label{eq6}
\begin{aligned}
\boldsymbol{\Sigma_{\mathbf{n}, \mathbf{q}}}
& =\sum_{i=1}^N \frac{\partial \mathbf{f}}{\partial {{\mathbf{p}_i}}} \boldsymbol{\Sigma_{{\mathbf{p}_i}}} {\frac{\partial \mathbf{f}}{\partial {{\mathbf{p}_i}}}}^T \\
& =\sum_{i=1}^N
\left[\begin{array}{ll}
\frac{\partial \mathbf{n}}{\partial {{\mathbf{p}_i}}} \boldsymbol{\Sigma_{{\mathbf{p}_i}}} {\frac{\partial \mathbf{n}}{\partial {{\mathbf{p}_i}}}}^T
&\frac{\partial \mathbf{n}}{\partial {{\mathbf{p}_i}}} \boldsymbol{\Sigma_{{\mathbf{p}_i}}} {\frac{\partial \mathbf{q}}{\partial {{\mathbf{p}_i}}}}^T \\
\frac{\partial \mathbf{q}}{\partial {{\mathbf{p}_i}}} \boldsymbol{\Sigma_{{\mathbf{p}_i}}} {\frac{\partial \mathbf{n}}{\partial {{\mathbf{p}_i}}}}^T
& \frac{\partial \mathbf{q}}{\partial {{\mathbf{p}_i}}} \boldsymbol{\Sigma_{{\mathbf{p}_i}}} {\frac{\partial \mathbf{q}}{\partial {{\mathbf{p}_i}}}}^T
\end{array}\right] \\
& =\left[\begin{array}{ll}
\sum_{i=1}^N \frac{\partial \mathbf{n}}{\partial {{\mathbf{p}_i}}} \boldsymbol{\Sigma_{{\mathbf{p}_i}}} {\frac{\partial \mathbf{n}}{\partial {{\mathbf{p}_i}}}}^T
& \sum_{i=1}^N \frac{\partial \mathbf{n}}{\partial {{\mathbf{p}_i}}} \boldsymbol{\Sigma_{{\mathbf{p}_i}}} {\frac{\partial \mathbf{q}}{\partial {{\mathbf{p}_i}}}}^T \\
\sum_{i=1}^N \frac{\partial \mathbf{q}}{\partial {{\mathbf{p}_i}}} \boldsymbol{\Sigma_{{\mathbf{p}_i}}} {\frac{\partial \mathbf{n}}{\partial {{\mathbf{p}_i}}}}^T
& \sum_{i=1}^N \frac{\partial \mathbf{q}}{\partial {{\mathbf{p}_i}}} \boldsymbol{\Sigma_{{\mathbf{p}_i}}} {\frac{\partial \mathbf{q}}{\partial {{\mathbf{p}_i}}}}^T
\end{array}\right] \\
& \stackrel{\operatorname{def}}{=}
\left[\begin{array}{cc}
\boldsymbol{\Sigma_{\mathbf{n} \mathbf{n}}} & \boldsymbol{\Sigma_{\mathbf{n} \mathbf{q}}} \\
\boldsymbol{\Sigma_{\mathbf{n} \mathbf{q}}}^T & \boldsymbol{\Sigma_{\mathbf{q} \mathbf{q}}}
\end{array}\right]
\end{aligned}
\end{equation}

Given the similarity in form among $\boldsymbol{\Sigma_{\mathbf{n} \mathbf{n}}}$, $\boldsymbol{\Sigma_{\mathbf{n} \mathbf{q}}}$, and $\boldsymbol{\Sigma_{\mathbf{q} \mathbf{q}}}$, we will use $\boldsymbol{\Sigma_{\mathbf{n} \mathbf{n}}}$ as an example to provide a detailed derivation of the cumulative update process for plane covariance. 
First, we substitute  (\ref{eq:partial}) in $\boldsymbol{\Sigma_{\mathbf{n} \mathbf{n}}}$  
\begin{equation}
\begin{aligned}
\label{eq7}
\boldsymbol{\Sigma_{\mathbf{n} \mathbf{n}}}
& =\sum_{i=1}^N \frac{\partial \mathbf{n}}{\partial {{\mathbf{p}_i}}} \boldsymbol{\Sigma_{{\mathbf{p}_i}}} {\frac{\partial \mathbf{n}}{\partial {{\mathbf{p}_i}}}}^T \\
&= \sum_{i=1}^N\mathbf{U}\left[\begin{array}{c}
\left( \mathbf{p}_i-\mathbf{q}\right)^T\mathbf{F}_{1} \\
\left( \mathbf{p}_i-\mathbf{q}\right)^T\mathbf{F}_{2} \\
\left( \mathbf{p}_i-\mathbf{q}\right)^T\mathbf{F}_{3}
\end{array}\right]\boldsymbol{\Sigma_{{\mathbf{p}_i}}}\left[\begin{array}{c}
\mathbf{F}_{1}^T\left( \mathbf{p}_i-\mathbf{q}\right) \\
\mathbf{F}_{2}^T\left( \mathbf{p}_i-\mathbf{q}\right) \\
\mathbf{F}_{3}^T\left( \mathbf{p}_i-\mathbf{q}\right)
\end{array}\right]\mathbf{U}^T \\
&= \mathbf{U}\sum_{i=1}^N\left[\begin{array}{c}
\left( \mathbf{p}_i-\mathbf{q}\right)^T\mathbf{F}_{1} \\
\left( \mathbf{p}_i-\mathbf{q}\right)^T\mathbf{F}_{2} \\
\left( \mathbf{p}_i-\mathbf{q}\right)^T\mathbf{F}_{3}
\end{array}\right]\boldsymbol{\Sigma_{{\mathbf{p}_i}}}\left[\begin{array}{c}
\mathbf{F}_{1}^T\left( \mathbf{p}_i-\mathbf{q}\right) \\
\mathbf{F}_{2}^T\left( \mathbf{p}_i-\mathbf{q}\right) \\
\mathbf{F}_{3}^T\left( \mathbf{p}_i-\mathbf{q}\right)
\end{array}\right]\mathbf{U}^T \\
& \stackrel{\operatorname{def}}{=} \mathbf{U}
\mathbf{B}  \mathbf{U}^T
\end{aligned}
\end{equation}
where $\mathbf{B}\in\mathbb{R}^{3\times3}$.

This transformation isolates $\mathbf{U}$ from the point-related computation.
Thus, to derive an update formula concerning new points, we only need to compute $\mathbf{B}$ cumulatively.

Let $b_{mn}$ be the element at $m$-th row and $n$-th column of $\mathbf{B}$.
\begin{equation}
\begin{aligned}
\label{eq9}
b_{mn}
&=\sum_{i=1}^N{\left({{\mathbf{p}_i}}-\mathbf{q}\right)^T \mathbf{\mathbf{F}_m} \boldsymbol{\Sigma_{{\mathbf{p}_i}}} \mathbf{\mathbf{F}_n}^T\left({{\mathbf{p}_i}}-\mathbf{q}\right)} \\
&=\sum_{i=1}^N {{{\mathbf{p}_i}}}^T \mathbf{F}_m \boldsymbol{\Sigma_{{\mathbf{p}_i}}} \mathbf{F}_n^T {{\mathbf{p}_i}}
-\sum_{i=1}^N {{{\mathbf{p}_i}}}^T \mathbf{F}_m \boldsymbol{\Sigma_{{\mathbf{p}_i}}} \mathbf{F}_n^T \mathbf{q}\\
&-\sum_{i=1}^N \mathbf{q}^T \mathbf{F}_m \boldsymbol{\Sigma_{{\mathbf{p}_i}}} \mathbf{F}_n^T {{{\mathbf{p}_i}}} +\sum_{i=1}^N \mathbf{q}^T \mathbf{F}_m \boldsymbol{\Sigma_{{\mathbf{p}_i}}} \mathbf{F}_n^T \mathbf{q}
\end{aligned}
\end{equation}

We would like to leverage statistics of all points to get rid of the dependency on the value of each point individually.
To achieve this goal, we introduce two lemmas below.
\begin{lemma}
\label{lemma:1}
Given a standard orthogonal basis in $\mathbb{R}^3$:  $\mathbf{e}_1=\left[1,0,0\right]^T, \mathbf{e}_2=\left[0,1,0\right]^T, \mathbf{e}_3=\left[0,0,1\right]^T$
$\forall \mathbf{A} \in \mathbb{R}^{3 \times 3}$ , each element of $\mathbf{A}$ can be represented as $\mathbf{a}_{ij}=\mathbf{e}_i^T\mathbf{A}\mathbf{e}_{j}$ ,where $i,j\in\{1,2,3\}$
\end{lemma}

\begin{lemma}
\label{lemma:2}
Given $\mathbf{x}$, $\mathbf{y} \in \mathbb{R}^3$, $\mathbf{A} \in \mathbb{R}^{3 \times 3}$, then $\mathbf{x}^T\mathbf{A}\mathbf{y} \in \mathbb{R}$, $\mathbf{y}\mathbf{x}^T\mathbf{A} \in \mathbb{R}^{3 \times 3}$ and $\mathbf{x}^T\mathbf{A}\mathbf{y}=\operatorname{tr}(\mathbf{x}^T\mathbf{A}\mathbf{y})=\operatorname{tr}(\mathbf{y}\mathbf{x}^T\mathbf{A})$, where $\operatorname{tr}(.)$  is the trace of the matrix.
\end{lemma}

Let $b_{mn}^{pp}$ be the first term of $b_{mn}$ in  (\ref{eq9}), i.e.,

\begin{center}
$b_{mn}^{pp}=\sum_{i=1}^N {{{\mathbf{p}_i}}}^T \mathbf{F}_m \boldsymbol{\Sigma_{{\mathbf{p}_i}}} \mathbf{F}_n^T {{\mathbf{p}_i}}$
\end{center}

Below we show how to update $b_{mn}^{pp}$ cumulatively.
Other terms of $b_{mn}$ can be computed similarly.

Let $\mathbf{v}_i^{m}=\mathbf{F}_m^T{{\mathbf{p}_i}}\in\mathbb{R}^{3 \times 1}$ , $\mathbf{D}_{i}=\boldsymbol{\Sigma_{{\mathbf{p}_i}}}\in\mathbb{R}^{3\times3}$, $\mathbf{v}_{i}^{n}=\mathbf{F}_n^T{{\mathbf{p}_i}}\in\mathbb{R}^{3\times1}$.

According to Lemma 1, each element of $\mathbf{v}_{i}^{m}$, $\mathbf{v}_{i}^{n}$ and $\mathbf{D}_{i}$ can be explicitly represented by the basis.
\begin{align}
\begin{split}
\label{eq10}
\mathbf{v}_i^{m} & =\left[
\begin{array}{c}
\mathbf{e}_1^T \mathbf{F}_m^T {{\mathbf{p}_i}}\\
\mathbf{e}_2^T \mathbf{F}_m^T {{\mathbf{p}_i}}\\
\mathbf{e}_3^T \mathbf{F}_m^T {{\mathbf{p}_i}}
\end{array}\right]  \qquad \mathbf{v}_{i}^{n} = \left[
\begin{array}{c}
\mathbf{e}_1^T \mathbf{F}_n^T {{\mathbf{p}_i}}\\
\mathbf{e}_2^T \mathbf{F}_n^T {{\mathbf{p}_i}}\\
\mathbf{e}_3^T \mathbf{F}_n^T {{\mathbf{p}_i}}
\end{array}\right]
\end{split}\\
\begin{split}
\label{eq11}
\mathbf{D}_{i} &= \left[\begin{array}{lll}
\mathbf{e}_1^T \boldsymbol{\Sigma_{{\mathbf{p}_i}}} \mathbf{e}_1 & \mathbf{e}_1^T \boldsymbol{\Sigma_{{\mathbf{p}_i}}} \mathbf{e}_2 & \mathbf{e}_1^T \boldsymbol{\Sigma_{{\mathbf{p}_i}}} \mathbf{e}_3 \\
\mathbf{e}_2^T \boldsymbol{\Sigma_{{\mathbf{p}_i}}}\mathbf{e}_1 & \mathbf{e}_2^T \boldsymbol{\Sigma_{{\mathbf{p}_i}}} \mathbf{e}_2 & \mathbf{e}_2^T \boldsymbol{\Sigma_{{\mathbf{p}_i}}} \mathbf{e}_3 \\
\mathbf{e}_3^T \boldsymbol{\Sigma_{{\mathbf{p}_i}}} \mathbf{e}_1 & \mathbf{e}_3^T \boldsymbol{\Sigma_{{\mathbf{p}_i}}} \mathbf{e}_2 & \mathbf{e}_3^T \boldsymbol{\Sigma_{{\mathbf{p}_i}}} \mathbf{e}_3
\end{array}\right]
\end{split}
\end{align}

Substituting  (\ref{eq10}-\ref{eq11}) in $b_{mn}^{pp}$ and expanding it yields a more concise form:
\begin{equation}
\label{eq13}
\begin{aligned}
b_{mn}^{pp}
&=\sum_{i=1}^N\mathbf{v}_i^{mT}\mathbf{D}_{i}\mathbf{v}_{i}^{n} \\
&=\sum_{i=1}^N \sum_{j=1}^3 \sum_{k=1}^3 {{\mathbf{p}_i}}^T \mathbf{F}_m \mathbf{e}_j \mathbf{e}_j^T \boldsymbol{\Sigma_{{\mathbf{p}_i}}} \mathbf{e}_k \mathbf{e}_k^T \mathbf{F}_n^T {{\mathbf{p}_i}}
\end{aligned} 
\end{equation}

According to the trace trick in Lemma 2 and the associative property of matrix multiplication,  (\ref{eq13}) can be further transformed as follows.
\begin{equation}
\begin{aligned}
b_{mn}^{pp}
=&\sum_{i=1}^N \sum_{j=1}^3 \sum_{k=1}^3 \operatorname{tr}\left({{\mathbf{p}_i}}^T \mathbf{F}_m \mathbf{e}_j \mathbf{e}_j^T \boldsymbol{\Sigma_{{\mathbf{p}_i}}} \mathbf{e}_k \mathbf{e}_k^T \mathbf{F}_n^T {{\mathbf{p}_i}}\right) \\
=&\sum_{i=1}^N \sum_{j=1}^3 \sum_{k=1}^3 \operatorname{tr}\left({{\mathbf{p}_i}} {{\mathbf{p}_i}}^T \mathbf{F}_m \mathbf{e}_j \mathbf{e}_j^T \boldsymbol{\Sigma_{{\mathbf{p}_i}}} \mathbf{e}_k \mathbf{e}_k^T \mathbf{F}_n^T\right) \\
=&\sum_{i=1}^N \sum_{j=1}^3 \sum_{k=1}^3 \operatorname{tr}\left(\mathbf{e}_j^T \boldsymbol{\Sigma_{{\mathbf{p}_i}}} \mathbf{e}_k {{\mathbf{p}_i}} {{\mathbf{p}_i}}^T \mathbf{F}_m \mathbf{e}_j \mathbf{e}_k^T \mathbf{F}_n^T\right) \\
=&\sum_{j=1}^3 \sum_{k=1}^3 \operatorname{tr}\left(\sum_{i=1}^N \left(\mathbf{e}_j^T \boldsymbol{\Sigma_{{\mathbf{p}_i}}} \mathbf{e}_k {{\mathbf{p}_i}} {{\mathbf{p}_i}}^T \right) \mathbf{F}_m \mathbf{e}_j \mathbf{e}_k^T \mathbf{F}_n^T\right) \\
=&\sum_{j=1}^3 \sum_{k=1}^3 \operatorname{tr}\left(\mathbf{X}_{j,k} \mathbf{F}_m \mathbf{e}_j \mathbf{e}_k^T \mathbf{F}_n^T\right)
\end{aligned} 
\end{equation}
where
\begin{equation}
    \mathbf{X}_{j,k}=\sum_{i=1}^N \mathbf{e}_j^T \boldsymbol{\Sigma_{{\mathbf{p}_i}}} \mathbf{e}_k {{\mathbf{p}_i}} {{\mathbf{p}_i}}^T \quad \forall j,k\in\{1,2,3\}
\end{equation}

It is noteworthy that terms related to points, specifically $\mathbf{p}_i$ and its covariance matrix $\boldsymbol{\Sigma_{{\mathbf{p}_i}}}$, can be incrementally accumulated in $\mathbf{X}_{j,k}$, thanks to $\mathbf{X}_{j,k}$ being independent of $\mathbf{F}_m$. Furthermore, in  (\ref{eq3-1}), $\mathbf{F}_m$ is computed from the singular vectors of the covariance matrix $\mathbf{A}$, which has a trivial cumulative update formula in  (\ref{eqA}).
Similarly, all terms in \Snq can be updated cumulatively with three statistics $\mathbf{X}_{j,k},\mathbf{Y}_j$ and $\mathbf{Z}$.
\begin{align}
\begin{split}
\label{sum2}
\mathbf{Y}_j &= \sum_{i=1}^N \boldsymbol{\Sigma_{{\mathbf{p}_i}}} \mathbf{e}_j {{\mathbf{p}_i}}^T \quad \qquad \forall j\in\{1,2,3\}
\end{split}\\
\begin{split}
\label{sum3}
\mathbf{Z} &= \sum_{i=1}^N \boldsymbol{\Sigma_{{\mathbf{p}_i}}}
\end{split}
\end{align}

Given a new point $\mathbf{p}_i$ and its covariance matrix $\boldsymbol{\Sigma_{\mathbf{p}_i}}$, all three statistics can be updated cumulatively.

\subsection{Complexity Analysis}
\label{sec:complexity}

As $3 \times 3$ symmetric matrices, only the upper-triangles of $\mathbf{X}_{j,k}$ and $\mathbf{Z}$ need to be stored, each with six scalars respectively. 
Given that $\mathbf{Y}_j$ is asymmetric, nine scalars are needed. 
Considering the number of matrices for each statistic, we can store all statistic matrices for a plane's uncertainty in a total of $ 6 \times 6 + 3 \times 9 + 6 = 69$ scalars.

In contrast, non-cumulative methods like VoxelMap\cite{VOXELMAP} store each point in a 3D vector and a $3 \times 3$ covariance matrix, resulting in a space complexity of $O(N)$ where $N$ is the number of points.
Our compact representation has a constant $O(1)$ space complexity regardless of the number of points.

Due to the re-computation in plane uncertainty update, non-cumulative methods have a time complexity of $O(MN)$ where $M$ is the number of iterations.
Our cumulative method reduces the time complexity to $O(N)$, making the updates much more efficient and scalable.

\section{On-demand Voxel Merging}
\label{sec:on-demand-merging}

The cumulative update strategy reduces the computation and storage for each voxel.
Further improvement is achieved by reducing the number of voxels through merging.
To reduce the overhead of searching for coalescible voxels in a brute-force manner, we design a locality-sensitive hash to aggregate voxels aligning on similar physical planes (Section \ref{sec:LSH}).
We propose the merge procedure and criterion to ensure merging consistency  (Section \ref{sec:on-demand}).
Moreover, we analyze the accuracy improvement of plane estimation due to voxel merging (Section \ref{sec:accuracy_analysis}).

\begin{figure}[t]
    \centering
    \includegraphics[width=0.6\linewidth]{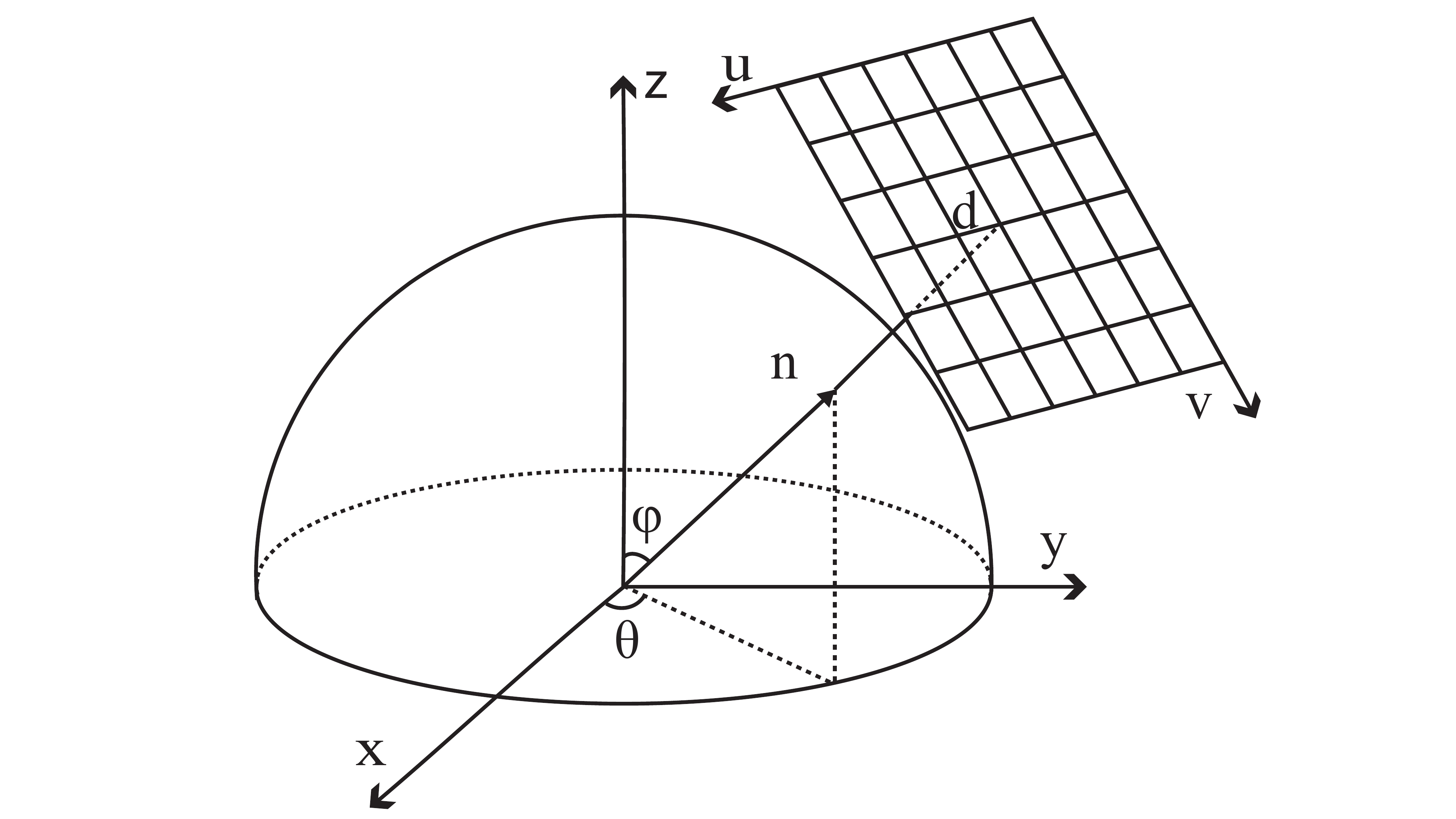}
    \caption{Locality-sensitive hash keys are comprised of five parameters: $\theta$, $\varphi$ are spherical coordinates of normal $\mathbf{n}$, $d$ is distance from origin to the plane, $u$ and $v$ are projection coordinates of a voxel on the plane.}
    \label{fig:lsh}
\end{figure}

\subsection{Locality-Sensitive Hash}
\label{sec:LSH}

To hash voxel planes according to their proximity in the parameter space,
we employ $[\theta, \varphi, d]$ for a representation with minimum DoF, where azimuthal angle $\theta$ and polar angle $\varphi$ are spherical coordinates computed from the normal vector $\mathbf{n}$, as in Fig.  \ref{fig:lsh}, and $d$ is computed by 
\begin{equation}
d=-\mathbf{n}^T\mathbf{q} 
\end{equation}
Given that a physical plane has limited area in the real world, voxel planes with similar parameters $\theta, \varphi, d$ but distributed far away might be associated with different real-world features and should not be merged.
To accommodate this property, we introduce a proximity coordinates $[u,v]$ computed as follows:
\begin{equation}
\left[\begin{array}{c}
u\\
v
\end{array}\right]=\left[
\begin{array}{c}
\mathbf{e}_1^T\mathbf{R}^{T}\mathbf{q}\\
\mathbf{e}_2^T\mathbf{R}^{T}\mathbf{q}
\end{array}
\right]
\end{equation}
where $\mathbf{R}$ indicates the rotation transform from global frame to the plane as computed in (\ref{Transform}), making $z$-axis aligns with the direction of $\mathbf{n}$, and $\mathbf{e}_1$, $\mathbf{e}_2$ are defined in \textit{Lemma} \ref{lemma:1}.
\begin{equation}
\label{Transform}
\mathbf{R}=\left[\begin{array}{ccc}
cos\theta & -sin\theta & 0 \\
sin\theta & cos\theta & 0 \\
0 & 0 & 1
\end{array}\right]
\left[\begin{array}{ccc}
cos\varphi & 0 & sin\varphi \\
0 & 1 & 0 \\
-sin\varphi & 0 & cos\varphi
\end{array}\right]
\end{equation}

A voxel plane is hashed into a bucket indexed by $\mathbf{k}\in\mathbb{R}^5$.
\begin{equation}
\label{eq:hashkey}
    \mathbf{k}=\left(\lfloor\frac{\theta}{\Delta\theta}\rfloor, \lfloor\frac{\varphi}{\Delta\varphi}\rfloor, \lfloor\frac{d}{\Delta d}\rfloor, \lfloor\frac{u}{\Delta u}\rfloor, \lfloor\frac{v}{\Delta v}\rfloor \right)^T
\end{equation}
where $\lfloor \cdot \rfloor$ rounds a number down to the nearest integer; $\Delta\theta$, $\Delta\varphi$, $\Delta d$, $\Delta u$ and $\Delta v$ are the widths of the bucket along each dimension.
Voxels with similar plane orientation and proximate to each other are likely to be hashed into the same bucket.

\subsection{Voxel Merging} 
\label{sec:on-demand}

\begin{figure}
    \centering
    \includegraphics[width=0.8\linewidth]{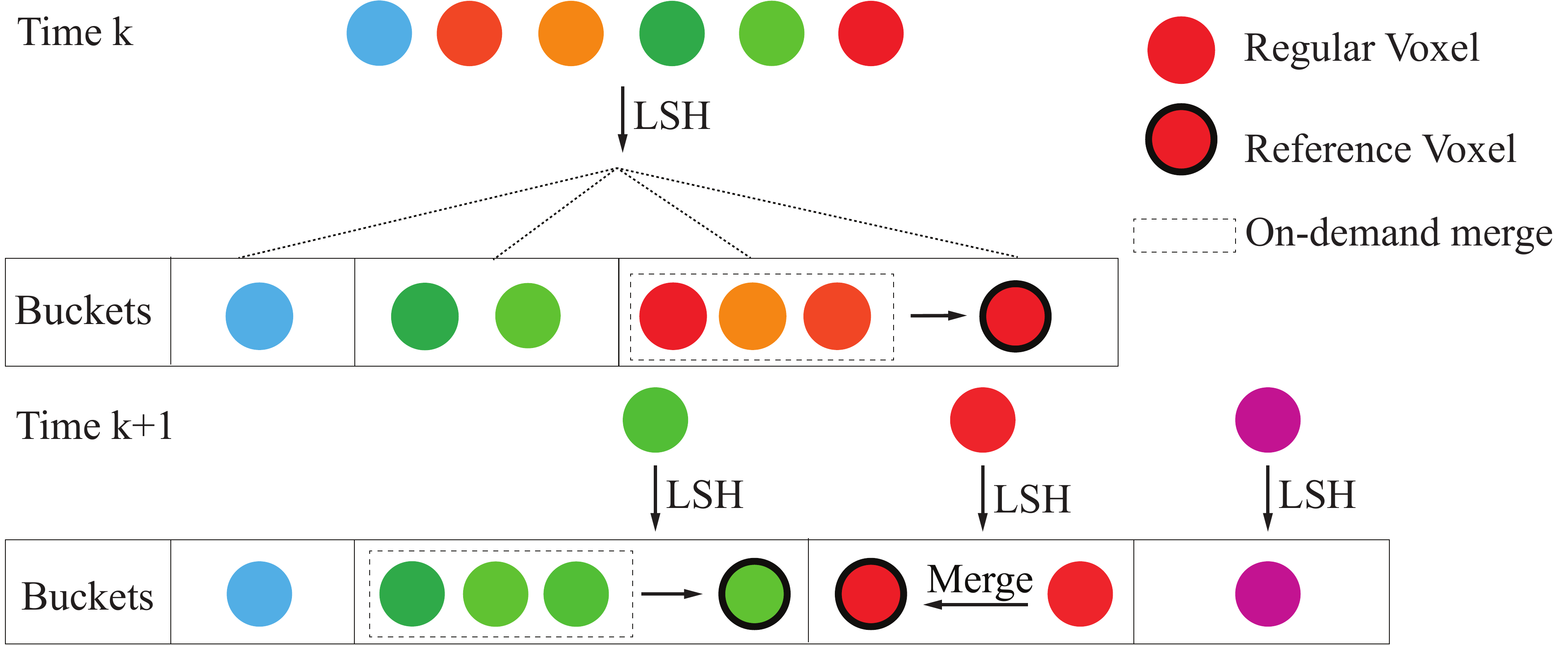}
    \caption{Illustration of voxel merging. Through LSH, voxel planes proximate to each other in the parameter space are hashed into the same bucket. A merge operation is triggered once a bucket accumulates enough voxels.}
    \label{fig:merge}
\end{figure}

When a hash bucket accumulates enough voxels, a merge operation is triggered as shown in Fig. \ref{fig:merge}.
Given multiple voxels in a bucket to be merged, we first find the voxel with the most points and then merge other voxels into this one by testing them against the merge criterion.
This is because the uncertainty of a plane is gradually reduced with point accumulation.
By taking the voxel with the most points as the reference, we are more confident in the merging correctness.

To avoid merging voxels associated with different physical planes by error, we check the following merging criterion.
Specifically, we require the eigenvalues of the covariance matrix $\mathbf{A}$ satisfy the planar assumption, i.e.,

\begin{equation}
\label{eigencheck}
    \frac{\lambda_3}{\lambda_1} < \eta \qquad \frac{\lambda_3}{\lambda_2} < \eta
\end{equation}
where $\lambda_1, \lambda_2, \lambda_3$ are eigenvalues of $\mathbf{A}$ in the descending order and $\eta$ is a sufficient small threshold closed to zero.

Once merged with the reference voxel of a bucket, the parameters of the merged voxels, i.e., $\mathbf{n}$, $\mathbf{q}$ and \Snq, as well as the memory allocated for them, are released. 
In contrast, the memory of a reference voxel is shared by other voxels merged into it, leading to a considerable saving of memory footprint.
Future points falling into any of these merged voxels cumulatively update the reference voxel.
Thanks to the cumulative update approach presented in Section \ref{Cumulative}, the statistics term of each voxel can be summed trivially without iterating the vast number of points for the large plane being merged. 

The voxel merging strategy enables a flexible representation of physical planes larger than the voxel dimension with only a single voxel.
Consequently, the challenge of bias-variance trade-off in terms of voxel size detailed in Section \ref{Introduction} can be addressed.

\subsection{Accuracy Analysis}
\label{sec:accuracy_analysis}

Voxel merging enhances mapping accuracy thanks to cross-voxel denoising.
We provide a theoretic analysis in this section.

Given two voxels with point set $\mathbb{P}$ and point set $\mathbb{Q}$ respectively. Both sets are sampled from the same physical plane. Suppose $\mathbb{P}$ and $\mathbb{Q}$ contains $M$ and $N$ points following normal distribution:
\begin{equation}
\mathbb{P}=\{\mathbf{p}_1, \mathbf{p}_2, ..., \mathbf{p}_M\} \quad \forall \mathbf{p}_i \sim \mathcal{N}(\boldsymbol{\mu}_p,\boldsymbol{\Sigma}_p)
\end{equation}
\begin{equation}
\mathbb{Q}=\{\mathbf{q}_1, \mathbf{q}_2, ..., \mathbf{q}_N\} \quad \forall \mathbf{q}_i \sim \mathcal{N}(\boldsymbol{\mu}_q,\boldsymbol{\Sigma}_q)
\end{equation}

According to Section \ref{Cumulative}, the mean $\boldsymbol{\mu}_{pq}$ and covariance matrix $\boldsymbol{\Sigma}_{pq}$ after merging $\mathbb{P}$ and $\mathbb{Q}$ can be updated cumulatively:
\begin{equation}
\boldsymbol{\mu}_{pq}=t\boldsymbol{\mu}_p+(1-t)\boldsymbol{\mu}_q, \quad t=\frac{M}{M+N}
\end{equation}
\begin{equation}
\label{eq:inc-update-sigma}
\boldsymbol{\Sigma}_{pq}=t\boldsymbol{\Sigma}_p + (1-t)\boldsymbol{\Sigma}_q + t(1-t)(\boldsymbol{\mu}_p-\boldsymbol{\mu}_q)(\boldsymbol{\mu}_p-\boldsymbol{\mu_q})^T
\end{equation}
where $t\boldsymbol{\Sigma}_p + (1-t)\boldsymbol{\Sigma}_q$ is linear combination of sample covariance weighted by points, and $\boldsymbol{\mu}_p-\boldsymbol{\mu}_q$ is a vector pointing from the center of $\mathbb{Q}$ to the center of $\mathbb{P}$. 
Here we introduce two lemmas to analyze why voxel merging benefits the accuracy. 

\begin{lemma}
\label{lemma:3}
Given a non-zero vector \(\mathbf{v} \in \mathbb{R}^n\), the matrix \(\mathbf{v}\mathbf{v}^T\) has only one non-zero eigenvalue \(\lambda = \lVert \mathbf{v} \rVert^2\) with the corresponding eigenvector \(\mathbf{v}\), where \(\lVert \cdot \rVert\) denotes the Euclidean norm. 
\end{lemma}

\begin{lemma}
\label{lemma:4}
For a symmetric matrix \(\mathbf{S}=\mathbf{v}\mathbf{v}^T\), where \(\mathbf{v}\) is a non-zero vector in \(\mathbb{R}^n\), the eigenvectors and singular vectors of \(\mathbf{S}\) are identical. The principal component of \(\mathbf{S}\) computed using Principal Component Analysis is in the direction of \(\mathbf{v}\).
\end{lemma}

\begin{figure}[h]
    \centering
    \includegraphics[width=1\linewidth]{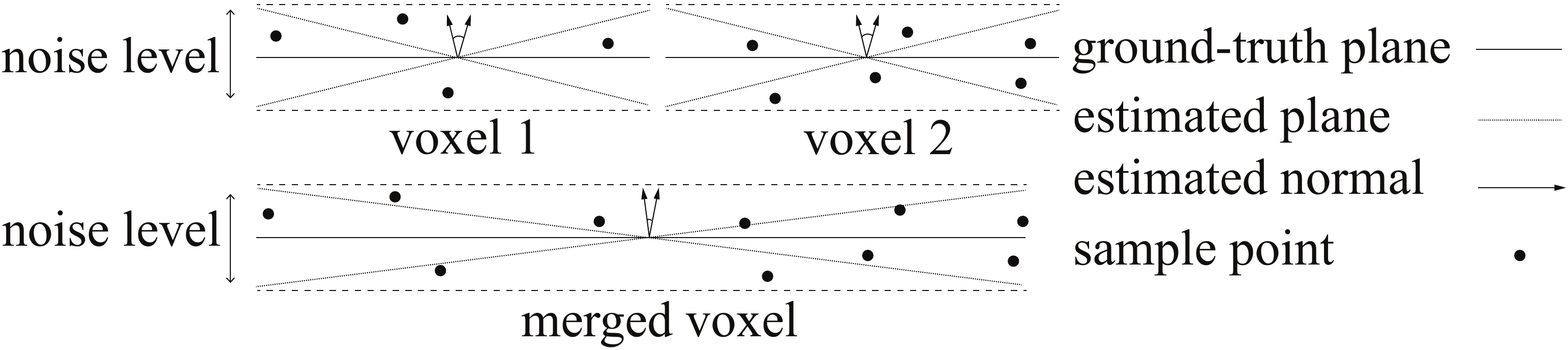}
    \caption{Illustration of accuracy enhancement thanks to voxel merging. The solid angle represents the robustness of plane estimation against noise. Merged voxel demonstrates a smaller angle than the original voxel under the same noise level.}
    \label{figure:merge_illustration}
\end{figure}

According to Lemma \ref{lemma:3} and \ref{lemma:4}, apart from the sample covariance $\boldsymbol{\Sigma}_p$ and $\boldsymbol{\Sigma}_q$, $\boldsymbol{\Sigma}_{pq}$ in  (\ref{eq:inc-update-sigma}) increases along the direction of $\boldsymbol{\mu}_p-\boldsymbol{\mu}_q$, which is expected to be on the physical plane.
The increased variance on the plane enhances the magnitudes of the first and second principal components.
As a result, the third principal component, i.e., the estimated normal direction in  (\ref{eq:normal}), gains increased robustness under the same noise level, as illustrated in Fig.  \ref{figure:merge_illustration}.

\section{Experiments}
\label{sec:experiments}
This section presents experiments on the accuracy and efficiency of C\textsuperscript{3}P-VoxelMap against the state-of-the-art on benchmarks including KITTI\cite{KITTI}, UTBM\cite{UTBM} and our self-collected data with a solid-state LiDAR Livox Mid360. 
The benchmarks cover both urban and indoor scenarios. 

We compare our method with five direct LO/LIO methods, namely Fast-LIO2\cite{FAST-LIO2}, Faster-LIO\cite{FASTER-LIO}, VoxelMap\cite{VOXELMAP}, VoxelMap++\cite{VOXELMAP++} and LiTAMIN2\cite{LITAMIN2}.
A comparison of these approaches is shown in Table \ref{method_comparison}. 
C\textsuperscript{3}P-VoxelMap is built on \cite{VOXELMAP} and publicly available on Github\footnote{\url{https://github.com/deptrum/c3p-voxelmap}}.

\subsection{Accuracy Evaluation}
\label{sec:Accuracy}
In this section, we evaluate our method on the odometry datasets of the KITTI Vision Benchmark\cite{KITTI} and UTBM\cite{UTBM}.
Default parameters are used to evaluate open-sourced algorithms without extra tuning. 
Specifically, for VoxelMap and our proposed method, the maximum voxel size is 3m and the maximum octo-tree layer is 3, leading to a minimum voxel size of 0.375m.
Since LiTAMIN2 is not open-sourced, we use results reported in the paper\cite{LITAMIN2}.
\begin{table}[h]
\centering
\caption{Accuracy (ATE In Meters) Comparison on UTBM Dateset}
\begin{tabular}{ccccccccccccc}
\hline
Approach & 0713 &0717 & 0719 & 0720  & Average \\ 
(Length [m]) & (5086) & (4998) & (4994) & (5035)  & (5028)  \\ \hline
Ours & \textbf{9.71} & \textbf{10.28} & \textbf{13.68} & \textbf{10.84}  & \textbf{11.09} & \\
VoxelMap & 10.99 & 11.26 & 15.80 & 14.58 & 13.13  &  \\
Faster-LIO & 14.88 & 15.36 & 16.30 & 15.05  & 15.38 & \\
Fast-LIO2 & 13.11 & 14.98 & 16.18 & 15.45 & 14.9 &  \\
\hline
\end{tabular}
\label{tab:utbm}
\end{table}
The results of experiments on UTBM and KITTI are listed in Table   \ref{tab:utbm} and \ref{tab:kitti_odometry}, respectively.
Since KITTI datasets provide no IMU data and the LiDAR scans have already been undistorted, a constant motion model is employed for all methods.
We utilize absolute trajectory error (ATE) defined in \cite{ATE} as the metric to evaluate all methods.
Lower ATE indicates higher odometry accuracy.
The last column is the average ATE of all sequences weighted by the number of frames.
It is shown that our method demonstrates around 20\% higher accuracy than the state-of-the-art \cite{VOXELMAP}, especially on long sequences like KITTI-02 and UTBM-0720.

From Table \ref{tab:utbm} and \ref{tab:kitti_odometry}, Fast-LIO2 and Faster-LIO are similar in terms of accuracy, because both methods employ the same deterministic plane representation and the same measurement model in state estimation. 
This experiment indicates that the voxel structure does not contribute directly to the accuracy compared with the k-d tree structure. 
In contrast, VoxelMap\cite{VOXELMAP} realizes higher accuracy by adopting probabilistic planes in the residual computation of IESKF.
Our method further improves the accuracy with respect to VoxelMap thanks to the proposed on-demand merging.

\begin{figure}[ht]
    \centering
    \includegraphics[width=1\linewidth]{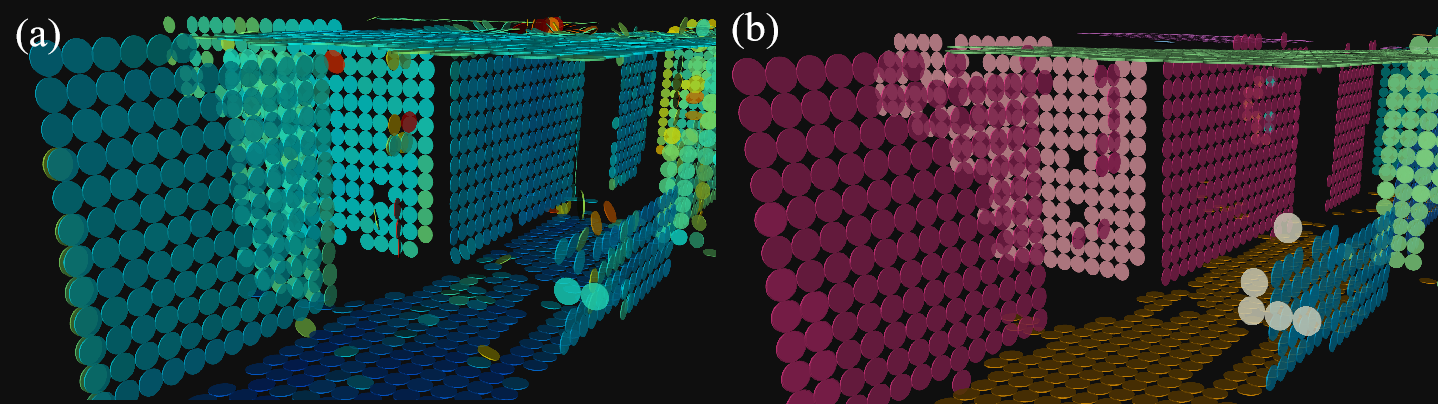}
    \caption{Illustration of voxel planes: (a) w/o merge and (b) w/ merge. Voxel merge eliminates the wiggling of planes and unifies the poses of planes associated with the same physical plane.}
    \label{fig:merge_result}
\end{figure}

For a better understanding of the accuracy improvement due to voxel merging, we visualize the merged voxels with distinct colors in Fig. \ref{fig:merge_result}.
The large plane features, e.g., ground, walls, and ceilings, are merged successfully.
Since the uncertainty of these planes is jointly determined by all merged voxels, a higher mapping accuracy is achieved than estimating the uncertainty of each voxel separately.

\begin{table*}[t]
\centering
\caption{Accuracy (ATE In Meters) Comparison on KITTI Odometry Training Sequences}
\begin{threeparttable}
\begin{tabular}{ccccccccccccc}
\hline
Approach & 00 &01 &02 &03 &04 &05 &06 &07 &08 &09 &10 & Average \\
(Length [m]) & (3724) & (2453) & (5067) & (560) & (393) & (2205) & (1232) & (694) & (3222) & (1705) & (919) &  (2016) \\ \hline
Ours & \textbf{1.97} & \textbf{3.69} & \textbf{5.74} & 0.92 & 0.26 & \textbf{0.98} & 0.31 & 0.56 & 3.33 & 2.65 & 1.34 & \textbf{2.74}\\
VoxelMap & 3.34 & 4.52 & 10.88 & 0.95 & \textbf{0.18} & 1.10 & \textbf{0.30} & 0.71 & 2.77 & 1.84 & 1.34 & 3.95 \\
VoxelMap++ & - & - & - & 1.39 & 0.41 & - & 0.33 & \textbf{0.55} & - & \textbf{1.40} & 1.71 & - \\
Faster-LIO & 5.016 & 19.495 & 9.692 & 0.977 & 0.397 & 1.661 & 2.142 & 1.018 & 4.307 & 2.326 & 1.866 & 5.25 \\
Fast-LIO2  & 9.426 & 20.77 & 9.537 & 1.06 & 0.435 & 1.999 & 2.152 & 1.15 & 4.361 & 2.458 & 2.399 & 6.24 \\
LiTAMIN2  & 5.8 & 15.9 & 10.7 & \textbf{0.8} & 0.7 & 2.4 & 0.9 & 0.6 & \textbf{2.5} & 2.1 & \textbf{1.0} & 5.1 \\
\hline
\end{tabular}
{\textbf{Bold} denotes the best accuracy for the case, ``-" indicates a significant drift from the ground-truth.}
\end{threeparttable}
\label{tab:kitti_odometry}
\end{table*}

\begin{table*}[h]
\centering
\caption{Time Evaluation (in milliseconds)}
\label{tab:time}
\begin{threeparttable}
\begin{tabular}{lSSSSSSSSSSSSSSSS}
\toprule
\multicolumn{2}{c}{\multirow{2}{*}{Map ID}} & \multicolumn{3}{c}{Ours (w/ merge)} & \multicolumn{3}{c}{Ours (w/o merge)} & \multicolumn{3}{c}{VoxelMap} & \multicolumn{3}{c}{Faster-LIO} & \multicolumn{3}{c}{Fast-LIO2}  \\ 
& & {E} & {M} & {Total} & {E} & {M} & {Total} & {E} & {M } & {Total} &{E } & {M} & {Total} &{E} & {M } & {Total}  \\ 
\midrule
\multirow{11}{*}{KITTI}& {00} & {23.54} & {7.25} & {30.79} & {26.06} & {4.36} & {30.42} & {26.06} & {14.01} & {40.07} &{47.49} & {1.01} & {48.50} & {35.30} & {3.43} & {38.73} \\
& {01} & {48.23} & {14.23} & {62.46} & {48.92} & {{9.41}} & {58.33} & {50.47} & {21.22} & {71.69} & {71.87} & {2.23} & {74.10} & {55.26} & {7.73} & {62.99} \\
& {02} & {25.35} & {8.05} & {33.40} & {24.26} & {{4.58}} & {28.84} & {25.84} & {16.36} & {42.20} & {46.81} & {1.10} & {47.91} & {35.07} & {3.77} & {38.84} \\
 & {03}& {35.60} & {10.40} & {46.00} & {35.72} & {{6.41}} & {42.13} & {37.41} & {25.38} & {62.79} & {60.05} & {1.29} & {61.34} & {45.58} & {4.05} & {49.63} \\
 & {04}& {38.28} & {11.05} & {49.33} & {37.24} & {{7.48}} & {44.72} & {38.82} & {22.55} & {61.37} & {56.96} & {1.58} & {58.54} & {50.23} & {6.19} & {56.42} \\
 & {05}& {31.52} & {8.66} & {40.18} & {31.15} & {{5.01}} & {36.16} & {33.57} & {16.88} & {50.45} & {55.53} & {0.97} & {56.50} & {37.14} & {3.16} & {40.30} \\
 & {06}& {46.83} & {12.90} & {59.73} & {45.09} & {{7.36}} & {52.45} & {47.97} & {29.99} & {77.96} & {89.70} & {1.32} & {91.02} & {55.77} & {3.59} & {59.36} \\
& {07} & {24.11} & {6.72} & {30.83} & {23.85} & {{4.16}} & {28.01} & {25.04} & {14.72} & {39.76} & {41.10} & {0.76} & {41.86} & {32.25} & {2.85} & {35.10} \\
& {08}& {39.32} & {10.66} & {49.98} & {39.31} & {{6.15}} & {45.46} & {42.06} & {18.91} & {60.97} & {69.34} & {1.39} & {70.73} & {45.23} & {4.28} & {49.51} \\
& {09} & {35.85} & {9.90} & {45.75} & {36.86} & {{6.15}} & {43.01} & {38.53} & {18.98} & {57.51} & {54.85} & {1.25} & {56.10} & {42.77} & {4.67} & {47.44} \\
& {10} & {23.74} & {6.77} & {30.51} & {23.70} & {{4.11}} & {27.81} & {24.50} & {14.20} & {38.7} & {38.39} & {0.81} & {39.20} & {30.39} & {2.92} & {33.31} \\
\midrule
& {Avg.} & {31.37} & {9.06} & {40.43} & {31.59} & {{5.38}} & {36.97} & {33.18} & {17.66} & {50.84} & {55.58} & {1.17} & {56.75} & {39.78} & {3.91} & {43.69} \\
\bottomrule
\end{tabular}
{ ``E" denotes the state estimation stage, ``M" denotes the map update stage.}
\end{threeparttable}

\end{table*}
\subsection{Efficiency Evaluation}
Besides accuracy, we evaluate both the computation performance and the memory consumption of our algorithm against others.
Experiments are conducted on a computer with Intel Core i5-12500H of 3.3GHz and 16GB RAM. 

We notice some open-source implementations employ multi-threading techniques, e.g., TBB in Faster-LIO,  OpenMP in Fast-LIO2 and VoxelMap.
To compare the efficiency of the algorithm itself, we leave out the implementation-related performance gain from multi-threading.
Thus, for a fair comparison, tests are performed on a single-threaded setup.

\textit{Time Efficiency}:
The performance results are listed in Table \ref{tab:time}.
All methods include two primary modules: state estimation and map update, and our C\textsuperscript{3}P-VoxelMap outperforms others in the overall processing time.

For the state estimation stage, although the same IESKF framework is employed for all the tested methods, the difference in processing time cannot be ignored. 
As listed in Table \ref{tab:time}, Faster-LIO consumes the longest 55.58 milliseconds in this stage as it requires a nearby voxel search which is time-consuming.
Fast-LIO2 also spends 8.41 milliseconds more time than ours in indexing neighbor points in the k-d tree and re-estimate planes.
In contrast, \cite{VOXELMAP} and our C\textsuperscript{3}P-VoxelMap obviate the need for NNS because each voxel contains an individual plane, which can be directly hashed in the voxel map.
Thus, our method saves 20\% and 43\% time usage compared with Fast-LIO2 and Faster-LIO, respectively. 

For the map update stage, thanks to our cumulative probabilistic update of voxel planes, the iteration of points and recalculation of Jacobians are eliminated and the time complexity is independent of the number of points. 
As a result, our C\textsuperscript{3}P-VoxelMap runs 70\%(w/o merge) and 49\%(w/ merge) faster than the original VoxelMap as shown in Table \ref{tab:time}. 
Since we not only update plane parameters but also update the uncertainty of the probabilistic plane, the time usage in our method takes longer than Fast-LIO2. 
However, our total processing time is 20\% lower than VoxelMap and 7\% lower than Fast-LIO2.

\textit{Memory Efficiency}:
We conduct two experiments to demonstrate the memory efficiency of our method against the others: First, we verify the constant complexity of cumulative probabilistic update. 
Second, we test the memory usage in both structured and unstructured environments. 

To verify the constant space complexity against the number of voxel points therein, we adjust the maximum number of points $N$ stored in each voxel when comparing our method with the original VoxelMap\cite{VOXELMAP}. 
If a voxel is updated by more than $N$ points, the plane uncertainty will not be updated anymore. 
We compare the memory consumption between the original VoxelMap and ours on KITTI's 06 sequence under different voxel granularity: a single-layer voxel map with a fixed 2m voxel size, and a multi-layer voxel map with 3m root voxel and a 3-layer sub-voxel as used in \ref{sec:Accuracy}.

\begin{figure}[h]
    \centering
    \includegraphics[width=0.8\linewidth]{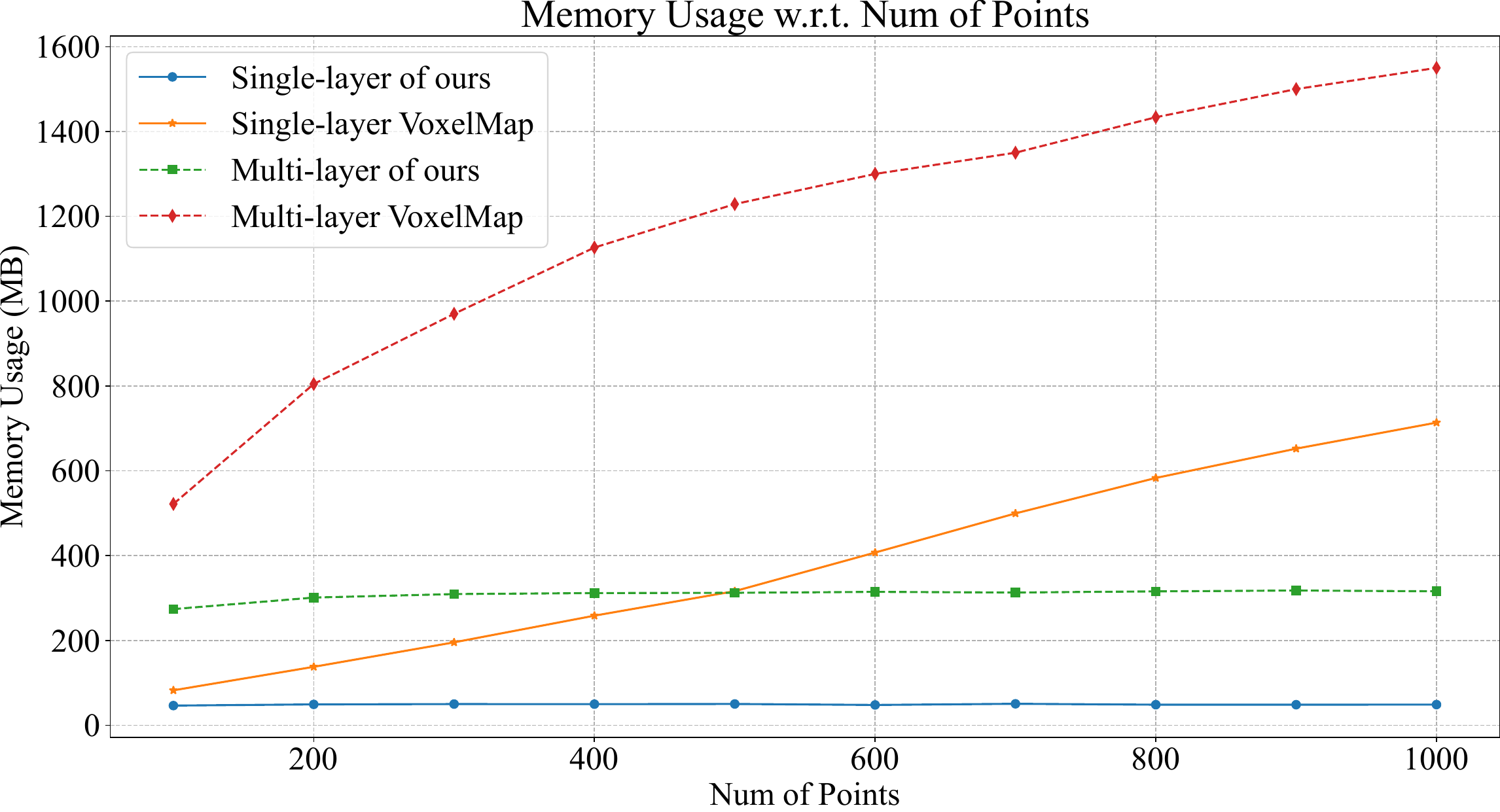}
    \caption{Comparison of memory usage between VoxelMap and ours.}
    \label{fig:memory}
\end{figure}
As illustrated in Fig. \ref{fig:memory}, the multi-layer experiment consumes more memory than the single-layer setup as the granularity of the map is finer. 
However, compared to the linear complexity of the original VoxelMap, our method only occupies a constant memory that is independent of the number of points used by the voxel map, leading to constant memory usage.

\begin{table}[h]
\centering
\caption{Comparison of Memory Usage (in mb)}
\begin{tabular}{ccccccccccccc}
\toprule
Approach & 04 & 06 & 09  & Indoor \\
(Length [m]) & (393) & (1232) & (1705) &  (560) \\ 
\midrule
C\textsuperscript{3}P-VoxelMap & 243.7 & 315.9 & 878.7  & 229.6 & \\
VoxelMap  & 521.0 & 1433.6 & 2150.4  & 1433.6 &  \\
\bottomrule
\end{tabular}
\label{tab:memory}
\end{table}
 
The comparison of the memory usage between C\textsuperscript{3}P-VoxelMap and the original VoxelMap is listed in Table \ref{tab:memory}. 
Besides urban environment tests on benchmark datasets, we also test our algorithm in an indoor environment with Livox Mid-360 LiDAR.
The results demonstrate that our method behaves 70\% greater memory efficiency than the VoxelMap. 
\section{Conclusion}
This work presents a compact, cumulative, and coalescible probabilistic voxel mapping method termed C\textsuperscript{3}P-VoxelMap, which involves a point-free representation of voxel planes, a cumulative probabilistic update method, and an on-demand voxel merging strategy.
The plane uncertainty is represented by a fixed set of statistics regardless of the number of points and supports cumulative updates without re-computation.
Our compact formulation reduces the space complexity from $O(N)$ to $O(1)$ and the time complexity from $O(MN)$ to $O(N)$ with respect to the number of iterations and the number of points \textit{N}.
The memory efficiency is further optimized by merging voxels associated with the same physical planes.
Thanks to the proposed locality-sensitive hash, the merging process is triggered on-demand with a small overhead.
On-demand merging enhances the adaptability of our voxel map to represent the real world and achieves a balance in the bias-variance tradeoff.
Moreover, this merging method improves the localization accuracy by smoothing out noises thanks to the cross-voxel updates.
Experiments on KITTI and UTBM benchmark demonstrate that our C\textsuperscript{3}P-VoxelMap outperforms the state-of-the-art with 20\% higher accuracy, 20\% higher performance, and 70\% lower memory consumption.
The high performance and low memory footprint of C\textsuperscript{3}P-VoxelMap show promises in achieving real-time LiDAR SLAM on resource-constrained platforms.

\bibliographystyle{IEEEtran}
\bibliography{C3P_VoxelMap}

\end{document}


\title{C\textsuperscript{3}P-VoxelMap Appendix}

\author{
\thanks{}
\thanks{}}


\maketitle

In this appendix, we focus on the cumulative update of the covariance matrix \Snq of the probabilistic plane:

\begin{equation}
\begin{aligned}
\nonumber
\boldsymbol{\Sigma_{\mathbf{n}, \mathbf{q}}}
=\left[\begin{array}{cc}
\boldsymbol{\Sigma_{\mathbf{n} \mathbf{n}}} & \boldsymbol{\Sigma_{\mathbf{n} \mathbf{q}}} \\
\boldsymbol{\Sigma_{\mathbf{n} \mathbf{q}}}^T & \boldsymbol{\Sigma_{\mathbf{q} \mathbf{q}}}
\end{array}\right]
\end{aligned}
\end{equation}

\Snq is composed of $\boldsymbol{\Sigma_{\mathbf{n} \mathbf{n}}}$, $\boldsymbol{\Sigma_{\mathbf{n} \mathbf{q}}}$ and $\boldsymbol{\Sigma_{\mathbf{q} \mathbf{q}}}$.
In the paper, the computation of $\boldsymbol{\Sigma_{\mathbf{n} \mathbf{n}}}$ has been partially discussed.
$\boldsymbol{\Sigma_{\mathbf{n} \mathbf{n}}}$ could be expressed by $\mathbf{U} \mathbf{B} \mathbf{U}^T$ where $\mathbf{U}$ is defined in Equation (3) of paper.
The element at $m$-th row and $n$-th column of $\mathbf{B}$ is denoted by $b_{mn}$, which is expressed as follows:

\begin{equation}
\begin{aligned}
\nonumber
b_{mn} &= b_{mn}^{pp} - b_{mn}^{pq} - b_{mn}^{qp} + b_{mn}^{qq} \\
b_{mn}^{pp} &= \sum_{i=1}^N {{{\mathbf{p}_i}}}^T \mathbf{F}_m \boldsymbol{\Sigma_{{\mathbf{p}_i}}} \mathbf{F}_n^T {{\mathbf{p}_i}} \\
b_{mn}^{pq} &= \sum_{i=1}^N {{{\mathbf{p}_i}}}^T \mathbf{F}_m \boldsymbol{\Sigma_{{\mathbf{p}_i}}} \mathbf{F}_n^T {{\mathbf{q}}} \\
b_{mn}^{qp} &= \sum_{i=1}^N {{{\mathbf{q}}}}^T \mathbf{F}_m \boldsymbol{\Sigma_{{\mathbf{p}_i}}} \mathbf{F}_n^T {{\mathbf{p}_i}} \\
b_{mn}^{qq} &= \sum_{i=1}^N {{{\mathbf{q}}}}^T \mathbf{F}_m \boldsymbol{\Sigma_{{\mathbf{p}_i}}} \mathbf{F}_n^T {{\mathbf{q}}}
\end{aligned}
\end{equation}
where $\mathbf{p}$ is a point with covariance $\boldsymbol{\Sigma_{{\mathbf{p}}}}$ on the plane, $\mathbf{q}$ is plane center and $\mathbf{F}$ is defined in Equation (6) of paper.

The paper has presented the computation of $b_{mn}^{pp}$ in Equation (13)-(16).
Similarly, $b_{mn}^{pq}$, $b_{mn}^{qp}$ and $b_{mn}^{qq}$ can be computed:

\begin{equation}
\begin{aligned}
\nonumber
b_{mn}^{pq}
=&\sum_{i=1}^N {{{\mathbf{p}_i}}}^T \mathbf{F}_m \boldsymbol{\Sigma_{{\mathbf{p}_i}}} \mathbf{F}_n^T {{\mathbf{q}}} \\
=&\sum_{i=1}^N \sum_{j=1}^3 {{\mathbf{p}_i}}^T \mathbf{F}_m \boldsymbol{\Sigma_{{\mathbf{p}_i}}} \mathbf{e}_j \mathbf{e}_j^T \mathbf{F}_n^T {{\mathbf{q}}} \\
=&\sum_{i=1}^N \sum_{j=1}^3 \operatorname{tr}\left({{\mathbf{p}_i}}^T \mathbf{F}_m  \boldsymbol{\Sigma_{{\mathbf{p}_i}}} \mathbf{e}_j \mathbf{e}_j^T \mathbf{F}_n^T {\mathbf{q}}\right) \\
=&\sum_{i=1}^N \sum_{j=1}^3 \mathbf{e}_j^T \mathbf{F}_n^T {\mathbf{q}} \operatorname{tr}\left({{\mathbf{p}_i}}^T \mathbf{F}_m \boldsymbol{\Sigma_{{\mathbf{p}_i}}} \mathbf{e}_j\right) \\
=&\sum_{i=1}^N \sum_{j=1}^3 \mathbf{e}_j^T \mathbf{F}_n^T {\mathbf{q}} \operatorname{tr}\left(\boldsymbol{\Sigma_{{\mathbf{p}_i}}} \mathbf{e}_j {{\mathbf{p}_i}}^T \mathbf{F}_m\right) \\
=&\sum_{j=1}^3 \mathbf{e}_j^T \mathbf{F}_n^T {\mathbf{q}} \operatorname{tr}\left(\left(\sum_{i=1}^N \boldsymbol{\Sigma_{{\mathbf{p}_i}}} \mathbf{e}_j {{\mathbf{p}_i}}^T \right) \mathbf{F}_m\right) \\
=&\sum_{j=1}^3 \mathbf{e}_j^T \mathbf{F}_n^T {\mathbf{q}} \operatorname{tr}\left({\mathbf{Y}_j} \mathbf{F}_m\right)
\end{aligned} 
\end{equation}\

\begin{equation}
\begin{aligned}
\nonumber
b_{mn}^{qp}
=&\sum_{i=1}^N {{{\mathbf{q}}}}^T \mathbf{F}_m \boldsymbol{\Sigma_{{\mathbf{p}_i}}} \mathbf{F}_n^T {{\mathbf{p}}} \\
=&\sum_{i=1}^N \left({{{\mathbf{q}}}}^T \mathbf{F}_m \boldsymbol{\Sigma_{{\mathbf{p}_i}}} \mathbf{F}_n^T {{\mathbf{p}}}\right)^T \\
=&\sum_{i=1}^N {\mathbf{p}_i}^T \mathbf{F}_n \boldsymbol{\Sigma_{{\mathbf{p}_i}}} \mathbf{F}_m^T {\mathbf{q}} \\
=&b_{nm}^{pq} \\
=&\sum_{j=1}^3 \mathbf{e}_j^T \mathbf{F}_m^T {\mathbf{q}} \operatorname{tr}\left({\mathbf{Y}_j} \mathbf{F}_n\right)
\end{aligned} 
\end{equation}

\begin{equation}
\begin{aligned}
\nonumber
b_{mn}^{qq}
=&\sum_{i=1}^N {\mathbf{q}}^T \mathbf{F}_m  \boldsymbol{\Sigma_{{\mathbf{p}_i}}} \mathbf{F}_n^T {\mathbf{q}} \\
=&{\mathbf{q}}^T \mathbf{F}_m \left(\sum_{i=1}^N \boldsymbol{\Sigma_{{\mathbf{p}_i}}} \right) \mathbf{F}_n^T {\mathbf{q}} \\
=&{\mathbf{q}}^T \mathbf{F}_m {\mathbf{Z}} \mathbf{F}_n^T {\mathbf{q}}
\end{aligned} 
\end{equation}
where $\mathbf{e}_j$ belongs to standard orthogonal basis in $\mathbb{R}^3$ and $\mathbf{Y}_j$, $\mathbf{Z}$ are two parameters that can be cumulatively updated:

\begin{equation}
\nonumber
    {\mathbf{Y}_j}=\sum_{i=1}^N \boldsymbol{\Sigma_{{\mathbf{p}_i}}} \mathbf{e}_j {{\mathbf{p}_i}}^T \quad
    {\mathbf{Z}}=\sum_{i=1}^N \boldsymbol{\Sigma_{{\mathbf{p}_i}}}
\end{equation}

Note that $\mathbf{q}$ and $\mathbf{F}$ can be updated in a cumulative way, which has been discussed in IV-B of the paper.
Therefore, all of $b_{mn}^{pp}$, $b_{mn}^{pp}$, $b_{mn}^{pp}$ and $b_{mn}^{pp}$ can be cumulatively update. 
Hence, the cumulative update of $b_{mn}$ is realized, so as $\boldsymbol{\Sigma_{\mathbf{n} \mathbf{n}}}$.

Next, $\boldsymbol{\Sigma_{\mathbf{n} \mathbf{q}}}$, which is defined in Equation (10) of paper, can be further expanded by using Equation (5) of paper:

\begin{equation}
\begin{aligned}
\nonumber
\boldsymbol{\Sigma_{\mathbf{n} \mathbf{q}}}
& =\sum_{i=1}^N \frac{\partial \mathbf{n}}{\partial {{\mathbf{p}_i}}} \boldsymbol{\Sigma_{{\mathbf{p}_i}}} {\frac{\partial \mathbf{q}}{\partial {{\mathbf{p}_i}}}}^T \\
&= \sum_{i=1}^N \frac{1}{N} \mathbf{U}\left[\begin{array}{c}
\left( \mathbf{p}_i-\mathbf{q}\right)^T\mathbf{F}_{1}\boldsymbol{\Sigma_{{\mathbf{p}_i}}} \\
\left( \mathbf{p}_i-\mathbf{q}\right)^T\mathbf{F}_{2}\boldsymbol{\Sigma_{{\mathbf{p}_i}}} \\
\left( \mathbf{p}_i-\mathbf{q}\right)^T\mathbf{F}_{3}\boldsymbol{\Sigma_{{\mathbf{p}_i}}}
\end{array}\right]\\
&= \frac{1}{N} \mathbf{U}\left[\begin{array}{c}
\sum_{i=1}^N \left( \mathbf{p}_i-\mathbf{q}\right)^T\mathbf{F}_{1}\boldsymbol{\Sigma_{{\mathbf{p}_i}}} \\
\sum_{i=1}^N \left( \mathbf{p}_i-\mathbf{q}\right)^T\mathbf{F}_{2}\boldsymbol{\Sigma_{{\mathbf{p}_i}}} \\
\sum_{i=1}^N \left( \mathbf{p}_i-\mathbf{q}\right)^T\mathbf{F}_{3}\boldsymbol{\Sigma_{{\mathbf{p}_i}}}
\end{array}\right] \\
& \stackrel{\operatorname{def}}{=} \frac{1}{N} \mathbf{U} \mathbf{C}
\end{aligned}
\end{equation}

Let $c_{mn}$ be the element at the $m$-th row and $n$-th column of $\mathbf{C}$. It can be computed by:

\begin{equation}
\begin{aligned}
\nonumber
c_{mn}
&=\sum_{i=1}^N \left( \mathbf{p}_i-\mathbf{q}\right)^T\mathbf{F}_{m}\boldsymbol{\Sigma_{{\mathbf{p}_i}}} \mathbf{e}_n\\
&=\sum_{i=1}^N {\mathbf{p}_i}^T\mathbf{F}_{m}\boldsymbol{\Sigma_{{\mathbf{p}_i}}} \mathbf{e}_n - \sum_{i=1}^N {\mathbf{q}}^T\mathbf{F}_{m}\boldsymbol{\Sigma_{{\mathbf{p}_i}}} \mathbf{e}_n \\
&=\sum_{i=1}^N \operatorname{tr}\left({\mathbf{p}_i}^T\mathbf{F}_{m}\boldsymbol{\Sigma_{{\mathbf{p}_i}}} \mathbf{e}_n\right) - {\mathbf{q}}^T\mathbf{F}_{m}\left(\sum_{i=1}^N \boldsymbol{\Sigma_{{\mathbf{p}_i}}}\right) \mathbf{e}_n \\
&=\sum_{i=1}^N \operatorname{tr}\left(\boldsymbol{\Sigma_{{\mathbf{p}_i}}} \mathbf{e}_n{\mathbf{p}_i}^T\mathbf{F}_{m}\right) - {\mathbf{q}}^T\mathbf{F}_{m}{\mathbf{Z}} \mathbf{e}_n \\
&= \operatorname{tr}\left(\left(\sum_{i=1}^N \boldsymbol{\Sigma_{{\mathbf{p}_i}}} \mathbf{e}_n{\mathbf{p}_i}^T\right) \mathbf{F}_{m}\right) - {\mathbf{q}}^T\mathbf{F}_{m}{\mathbf{Z}} \mathbf{e}_n \\
&= \operatorname{tr}\left({\mathbf{Y}_n} \mathbf{F}_{m}\right) - {\mathbf{q}}^T\mathbf{F}_{m}{\mathbf{Z}} \mathbf{e}_n \\
\end{aligned}
\end{equation}
where ${\mathbf{Y}_n}$ and ${\mathbf{Z}}$ are parameters which can be cumulatively updated given in previous equations. Thus, $c_{mn}$, so as $\boldsymbol{\Sigma_{\mathbf{n} \mathbf{q}}}$, can cumulatively updated.

Finally, $\boldsymbol{\Sigma_{\mathbf{q} \mathbf{q}}}$ is obtained by:

\begin{equation}
\begin{aligned}
\nonumber
\boldsymbol{\Sigma_{\mathbf{q} \mathbf{q}}}
= \frac{1}{N^2} \sum_{i=1}^N \boldsymbol{\Sigma_{{\mathbf{p}_i}}}
= \frac{1}{N^2} {\mathbf{Z}} 
\end{aligned}
\end{equation}

So far, the cumulative update of $\boldsymbol{\Sigma_{\mathbf{n} \mathbf{n}}}$, $\boldsymbol{\Sigma_{\mathbf{n} \mathbf{q}}}$ and $\boldsymbol{\Sigma_{\mathbf{q} \mathbf{q}}}$ have all been accomplished.
Therefore, the covariance matrix of probabilistic plane \Snq can be obtained cumulatively.